\documentclass[USletter, 10 pt, conference]{ieeeconf}  
\IEEEoverridecommandlockouts

\usepackage{amsmath,amssymb,amsfonts}
\usepackage{graphicx}

\usepackage{multirow}

\usepackage{siunitx}
\usepackage{bm}   

\IEEEoverridecommandlockouts                             
\title{\LARGE \bf Comparison of neural network training strategies for the simulation of dynamical systems}

\author{Paul Strasser$^{1}$, Andreas Pfeffer$^{1}$, Jakob Weber$^{1}$, Markus Gurtner$^{1}$, Andreas Körner$^{2}$
\thanks{$^{1}$Center for Vision, Automation \& Control, AIT Austrian Institute of Technology GmbH, Vienna, Austria. 
                {\tt\small andreas.pfeffer@ait.ac.at}}
\thanks{$^{2}$ Institute of Analysis and Scientific Computing, TU Wien, Vienna, Austria}
}

\begin{document}

\maketitle
\thispagestyle{empty}  
\pagestyle{empty}

\begin{abstract}
Neural networks have become a widely adopted tool for modeling nonlinear dynamical systems from data. 
However, the choice of training strategy remains a key design decision, particularly for simulation tasks. 
This paper compares two predominant strategies: \textit{parallel} and \textit{series-parallel} training. 
The conducted empirical analysis spans five neural network architectures and two examples: a pneumatic valve test bench and an industrial robot benchmark. 
The study reveals that, even though \textit{series-parallel} training dominates current practice, \textit{parallel} training consistently yields better long-term prediction accuracy.
Additionally, this work clarifies the often inconsistent terminology in the literature and relate both strategies to concepts from system identification. 
The findings suggest that \textit{parallel} training should be considered the default training strategy for neural network-based simulation of dynamical systems.

\end{abstract}


\section{Introduction}

Modeling complex dynamical systems using neural networks is an active research area, with demonstrated utility in control systems~\cite{Saviolo2022, Gordon_2024} and robotic navigation~\cite{roth2025learnedperceptiveforwarddynamics}. 
These data-driven models offer flexibility and expressive power but introduce key design challenges, particularly in how the networks are trained. 
While significant progress has been made in architecture design and optimization, the choice of training strategy remains less clearly understood.
When modeling systems from measurement data, especially in the context of time series prediction, two training paradigms dominate: \textit{parallel} and \textit{series-parallel} training. 
The main distinction lies in whether the model uses previous measured ground-truth outputs (\textit{series-parallel}) or its own previous predictions (\textit{parallel}) as input during training. 
In this work both strategies are systematically compared across several neural network architectures using real-world measurements from a pneumatic valve test bench~\cite{Glueck2021} and a widely used industrial robot benchmark~\cite{Weigand2022}.   
The results show that \textit{parallel} training consistently achieves higher predictive accuracy over \textit{series-parallel} training, indicating it should be considered the default strategy for similar modeling tasks. 
Furthermore, this work seeks to clarify and standardize terminology related to these training strategies.  
To that end, their connections to other machine learning domains, such as reinforcement learning, and to the theory of system identification~\cite{Pillonetto2025} are highlighted.
 
The remainder of the paper is structured as follows. 
Section~\ref{sec:methods} first formalizes the neural network-based simulation task (Sec.~\ref{subsec:sim_dyn_sys}), then introduces the five network architectures under study (Sec.~\ref{subsec:NN_architectures}), explains the \textit{parallel} and \textit{series-parallel} training strategies together with their links to system identification (Sec.~\ref{subsec:training_strategies}), and concludes with a review of related work and terminology (Sec.~\ref{subsec:literature}).
Section~\ref{sec:experiments} provides details on the experimental setup, covering data collection procedure for the pneumatic valve system (Sec.~\ref{subsec:valve}) and the industrial robot dataset (Sec.~\ref{subsec:robot}). 
Section~\ref{sec:discussion} reports and analyzes the results for all architectures and training strategies. 
Finally, Section~\ref{sec:conclusion} summarizes the findings and outlines directions for future research.

\section{Background and Methodology} \label{sec:methods}
This section outlines the methodological foundation of this study. 
It defines the neural network modeling task for dynamical systems, introduces the network architectures considered, and presents the two main training strategies - \textit{parallel} and \textit{series-parallel}. 
The section concludes with a discussion of related work and terminology.

\subsection{Simulation of Dynamical Systems} \label{subsec:sim_dyn_sys}
The dynamical systems of interest can be described by a continuous-time state-space model of the form
\begin{equation}\label{eq:dynsys}
    \begin{aligned} 
        \frac{\text{d} \bm{x}}{\text{d} t} &= \bm{f}(\bm{x},\bm{u})\ , \\
        \bm{y} &= \bm{g}(\bm{x},\bm{u})\ ,\\
        \bm{x}\vert_{t=0} &=\bm{x}_0 \\
    \end{aligned}     
\end{equation}
where $\bm{f}$ represent the system dynamics, $\bm{x}$ the states, $\bm{u}$ the inputs, $\bm{g}$ the output function, $\bm{y}$ the vector of measurable outputs and the initial condition $\bm{x}_0$.
Sampling the system at a constant rate with sampling time $T_s$ yields a sequence of input-output pairs $\{(\bm{y}_i, \bm{u}_i)\}_{i=1}^N$, where $\bm{y}_i = \bm{y}(t_i)$, $\bm{u}_i=\bm{u}(t_i)$, and $t_i = iT_s$.
The objective is to predict the future system output $\bm{y}_{k+1}$ using a neural network $\mathcal{N}$ based on $L$ previous inputs and outputs (typically $L > 1$). 
This results in the prediction
\begin{align}
    \bm{\hat y}_{k+1} = \mathcal{N}(\left[ \bm{y}_{k-L}, \dots, \bm{y}_k, \bm{u}_{k-L}, \dots, \bm{u}_k\right])\ .
\end{align}
In the context of system identification, this approach is referred to as black-box modeling, as it does not rely on prior knowledge of the underlying system.
Such an approach is well-suited for systems with complex, non-linear dynamics - such as those considered in this study. 

\subsection{Network Architectures} \label{subsec:NN_architectures}

In this section, a variety of network architectures are presented to support broader conclusions about the differences between the training methods. 
Five representative architectures are selected: two feedforward models - the Multilayer Perceptron (MLP) and the Temporal Convolutional Network (TCN) - and three recurrent models.
While this selection is not exhaustive and does not capture the full landscape of emerging architectures (see, e.g., ~\cite{FananasAnaya2025, roth2025learnedperceptiveforwarddynamics, Geneva2022, Paniagua2024, rosta2025}), recent innovations, including architectural adaptations~\cite{Oveissi2024}, are referenced throughout the paper for context.

The MLP consists of multiple layers of linear transformations followed by a nonlinear activation function. 
Since it lacks a specific mechanism for modeling temporal dependencies, multiple consecutive time steps are concatenated into a single input vector, enabling the network to infer temporal patterns implicitly.

In contrast, recurrent networks are specifically designed to capture temporal dependencies. 
This is achieved by maintaining a hidden state that is updated at each time step based on the current input and the previous hidden state. 
A simple Recurrent Neural Network (RNN) is described by
\begin{equation}\label{eq:rnn}
    \begin{aligned}
    \bm{h}_t &= \sigma(W_1\bm{h}_{t-1}+W_2\bm{x}_{t-1}+\bm{b}_1) \\
   \bm{y}_t &= \sigma(W_3\bm{h}_{t}+\bm{b}_2)\ ,
\end{aligned} 
\end{equation}   
where $W_i$ and $b_i$ are weight matrices and biases, respectively, and $\sigma$ denotes a nonlinear activation function.

Among recurrent architectures, the Long Short-Term Memory (LSTM) network~\cite{hochreiter1997} is one of the most widely used. 
The Gated Recurrent Unit (GRU) offers a simplified alternative to the LSTM~\cite{cho2014learning}. While all recurrent architectures are capable of modeling time series, their respective strengths and weaknesses make them suitable for different tasks~\cite{Collins2017}.

The Temporal Convolutional neural network (TCN) has been shown to achieve performance comparable to recurrent models in sequence modeling tasks~\cite{Bai2018}. 
It uses one-dimensional dilated causal convolutions across multiple layers, effectively capturing long-range temporal dependencies.

Several of the evaluated architectures incorporate skip connections, a common architectural feature in deep learning models~\cite{he2015}.
In cases where $\bm{g}(\bm{x}, \bm{u}) \equiv \bm{x}$ in~\eqref{eq:dynsys}, i.e. $\bm{x} = \bm{y}$, the use of a skip-connection can also be interpreted as enabling the network to approximate a derivative-like signal, as discussed in~\cite{Legaard2023}.

\subsection{Training Strategies} \label{subsec:training_strategies}
\begin{figure}[h!]
    \centering
    \includegraphics[width=0.48\textwidth]{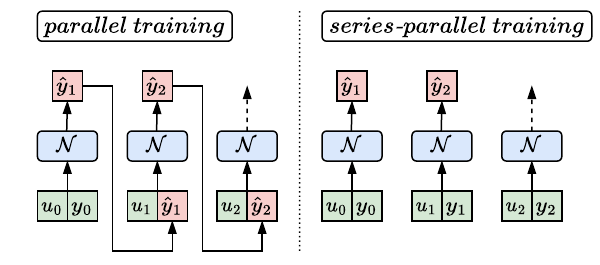}
    \caption{Information flows for \textit{parallel} training~\eqref{eq:parallel} and \textit{series-parallel} training~\eqref{eq:series-parallel}.}
    \label{fig:parallel_training}
\end{figure}

In \textit{series-parallel} training, the model predicts the next output based on the current input $\bm{u}_k$ and the measured system output $\bm{y}_k$, yielding
\begin{align}\label{eq:series-parallel}
    \hat{\bm{y}}_{k+1}  &= \mathcal{N}([\bm{y}_k, \bm{u}_k])\ .
\end{align}
In \textit{parallel} training, the prediction $\hat{\bm{y}}_k$ at the previous time step is used instead of the measured output $\bm{y}_k$, yielding
\begin{align}\label{eq:parallel}
    \hat{\bm{y}}_{k+1}  &= \mathcal{N}([\hat{\bm{y}}_k, \bm{u}_k])\ .
\end{align}
The different information flows, which affect how gradients are computed during training, are illustrated in Fig.~\ref{fig:parallel_training}.
While extensions~\cite{Ribeiro2017} and even combinations~\cite{Bengio2015} of these methods have been proposed, this work focuses on the basic formulations above. 
\textit{Series-parallel} training, as in~\eqref{eq:series-parallel}, offers efficient parallelization, enabling large batches of data to be processed simultaneously. 
In contrast, \textit{parallel} training~\eqref{eq:parallel} requires sequential computation, as each prediction depends on the previous one. 
Implementing~\eqref{eq:parallel} also requires preserving gradient paths through the predictions, for which in this work the well-established PyTorch library is used~\cite{Chen2022}.

\noindent

Once trained, both models can be deployed for forward simulation. 
This requires an initial system output and a sequence of system inputs. Predictions are generated recursively by feeding the model’s own outputs back as inputs - an approach commonly referred to as \textit{free-running simulation}. 
This method was used in the experiments described in Section~\ref{sec:experiments} allowing simulation over arbitrary horizons, as long as new inputs are available. 
While other inference modes, such as \textit{n-step-ahead-prediction} exist, they are not considered here, because they have not been used in the industrial robot benchmark~\cite{Weigand2022}.

\subsection{Literature Review} \label{subsec:literature}
The commonly held belief that \textit{series-parallel} training is the superior strategy - often attributed to early foundational work~\cite{Kumpati1990} - has been challenged in recent years. 
For instance, the study in~\cite{Ribeiro2018} compares both training methods with a focus on robustness, computational efficiency, and convergence behavior. 
In contrast, this paper examines how architectural choices influence the performance of these training strategies.
While these approaches are not new, their terminology remains inconsistent across the literature. 
Terms such as \textit{forecast training}~\cite{Noto2024}, LSTM networks with \textit{output recurrence}~\cite{Chen2022}, and \textit{training without teacher forcing}~\cite{Sangiorgio2020} are often used to describe what is referred to here as \textit{parallel} training.

From a system identification perspective, models trained using \textit{series-parallel} training are equivalent to nonlinear auto-regressive models with exogenous inputs (NARX), whereas models trained using \textit{parallel} training correspond to nonlinear output-error (NOE) models~\cite[Chapter~7]{Schroeder2017}. 
It is important to note that these model classes are general and not limited to neural networks. 
A more comprehensive discussion on the intersection of system identification and machine learning is provided in~\cite{Pillonetto2025}.

A known challenge in data-driven modeling is the presence of distribution shift between training and inference, which can lead to compounding errors over time.
In reinforcement learning, this issue is typically addressed through techniques such as exploration and online adaptation~\cite{wang2024, cho2024, ross2011}. 
Similarly, \textit{parallel} training helps to mitigate this problem by exposing the model to its own predictions during training, thereby reducing the mismatch between training and inference distributions.

\section{Experiments}\label{sec:experiments}
This chapter presents the experimental setup, including the systems under study, the training procedure, evaluation metrics, and implementation details.

\subsection{Pneumatic Valve} ~\label{subsec:valve}
The first system studied is a two-stage pneumatic valve, for which real-world measurements were collected using a test bench. 
This setup enabled controlled data acquisition with specified coverage of the system’s state space. 
A detailed mathematical model of the full system has been presented in~\cite{Glueck2021}; therefore, only a brief description of one of the four valves of one valve unit is provided here.

Each valve comprises a pre-stage and a main stage. 
The main stage consists of a plunger, preloaded on one side by a spring, with its position determined by the pressure in a control pressure chamber on the opposite side. 
This configuration effectively behaves as a single-acting cylinder with spring return. 
The directly measurable system outputs $\bm{y}$ are the plunger position $s$ and the control chamber pressure $p$. 
The system inputs $\bm{u}$ are two control voltages, each applied to a piezoelectrically actuated beam valve in the pre-stage. 
These voltages regulate the chamber pressure and, consequently, the plunger position. 
Together, the inputs and outputs form the feature vectors used for training: $[\bm{y}, \bm{u}]$.

The quality and diversity of the training data are critical for successful modeling. 
For dynamical systems, this means ensuring that a broad range of state-space trajectories are captured. 
To achieve this, a test signal generator described in~\cite{westhauser2025} was used to create diverse input sequences - comprising both static and dynamic sections - while respecting the physical limits of the system. 
The final training dataset includes 900 trajectories with 500 time steps each, sampled at \SI{40}{\micro\second}. 
The test set consists of 30 longer trajectories, each with 3500 time steps, thereby requiring the models to generalize across significantly longer horizons. The full dataset is publicly available~\cite{strasser_2025}.

\subsection{Industrial Robot} \label{subsec:robot}
The second system is an industrial robot, for which a benchmark dataset is available online~\cite{Weigand2022}. 
It includes 36 distinct trajectories, each executed twice to assess repeatability, totaling approximately 73 minutes of motion data. 
For the forward modeling task, the system inputs $\bm{u}$ are the motor torques $\tau_1, \dots, \tau_6$ at each of the six joints while the outputs $\bm{y}$ are the corresponding joint positions $q_1, \dots, q_6$.
Following preprocessing and resampling at \SI{100}{\milli\second}, the dataset was split into a training set with approx. 67 minutes and a test set with approx. six minutes of data. 
A linear state-space model is also provided as a baseline reference.

It is important to note that the amount of unique data is limited, particularly because each trajectory is repeated. 
While repetition helps mitigate system variability, it does not increase the effective state-space coverage. 
This limitation makes the training process more prone to overfitting, even with regularization techniques. 
In contrast, such issues were not encountered with the pneumatic valve dataset, where significantly more diverse training data was available. 
This comparison highlights the crucial impact of both the quantity and quality of data on the performance of learned models.

\subsection{Training}\label{sec:Training}
All models were implemented using the PyTorch framework and trained using the AdamW optimizer~\cite{kingma2017adammethodstochasticoptimization}, a variant of the Adam algorithm that incorporates decoupled weight decay. 
The optimization objective was the mean squared error (MSE) between predicted and measured outputs. 
An initial set of hyperparameters was determined for each architecture via a grid search using the Tune library~\cite{liaw2018tune}. 
To enhance training performance and mitigate overfitting, the following techniques were employed: L2 regularization, gradient clipping, early stopping, and dropout layers.

Consistent with the recommendation in~\cite{Weigand2022}, the normalized root mean squared error (NRMSE) was used to evaluate model performance by
\begin{align} \label{eq:nrmse}
    \text{NRMSE}(\bm{y},\hat{\bm{y}}) =  \sqrt{\frac{1}{n} \sum_{k=1}^{m}\sum_{i=1}^{n} \left(\frac{y^k_i - \hat{y}^k_i}{\sigma_k}\right )^2} \ ,
\end{align}
where $n$ is the number of simulation steps and $m$ is the number of outputs. 
The standard deviation was estimated by $\sigma_k=\sqrt{\frac{1}{n}\sum_{i=1}^n (y^k_i - \bar{y}^k)^2}$ with $\bar{y}^k$ as the mean of the $k$-th output over the prediction horizon. 

\subsection{Experiment Details}\label{sec:exp_details}
For the pneumatic valve experiment, 30 distinct trajectories were simulated using the test set inputs $\bm{u}_\text{test}$. 
The NRMSE, see~\eqref{eq:nrmse}, was computed across all test trajectories and then averaged. 
The diversity of initial conditions and the long prediction horizon provide a robust evaluation of each model’s generalization ability. 
While many related studies report results in terms of RMSE, using NRMSE consistently across both experiments facilitates direct comparison.

For a fair comparison, base hyperparameter configurations for each architecture were identified using the Tune library~\cite{liaw2018tune}. 
Afterwards, these were fine-tuned for both training strategies individually.  
While this setup improves comparability, it is not guaranteed that the conclusions generalize across all architectures or hyperparameter regimes.
For the industrial robot, the entire test set was concatenated into a single trajectory. 
The model was then evaluated on this continuous sequence to assess long-horizon prediction performance using the same error metrics described above.
Table~\ref{tab:nrmse_all} shows the average NRMSE for all architectures for the pneumatic valve and for the industrial robot benchmark.

\begin{figure}[!h]
    \centering
    \includegraphics[width=1\linewidth]{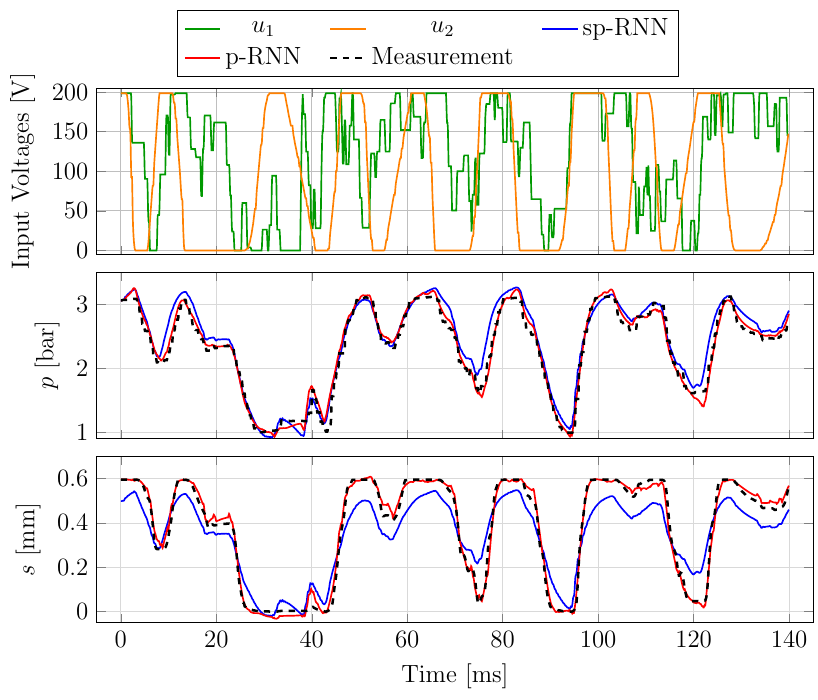}
    \caption{ 
    Predicted pressure $p$ and plunger position $s$ for the pneumatic valve system on one representative test trajectory. 
    The top two plots show the applied input voltages $u_1$ and $u_2$. 
    The bottom plots compare the outputs of the best-performing \textit{parallel}-trained model (p-RNN) and \textit{series-parallel}-trained model (sp-RNN) against measured data (black dashed lines).} 
    \label{fig:ventil_plots}
\end{figure}

\begin{table}[t]trial Robot base
  \centering
  \caption{Average NRMSE (\textit{lower is better}) for all network architectures, grouped by dataset and training strategy. The best performing networks are marked with a $*$. The industrial robot baseline from \cite{Weigand2022} achieves an NRMSE of 0.82.}
  \label{tab:nrmse_all}
  \renewcommand{\arraystretch}{1.1}
  \begin{tabular}{llcc}
    \hline\hline
    \textbf{Dataset} & \textbf{Network} & \textbf{Series-Parallel} & \textbf{Parallel} \\ \hline
    \multirow{5}{*}{Pneumatic Valve}
      & MLP  & 0.80 & \textbf{0.48} \\
      & RNN  & 0.64* & \textbf{0.22}* \\
      & LSTM & 0.82 & \textbf{0.32} \\
      & GRU  & 1.53 & \textbf{0.28} \\
      & TCN  & 1.33 & \textbf{0.33} \\ \hline
    \multirow{5}{*}{Industrial Robot}
      & RNN  & 1.51 & \textbf{0.77} \\
      & LSTM & 1.95 & \textbf{0.68} \\
      & GRU  & 1.76 & \textbf{0.67}* \\
      & MLP  & 1.94 & \textbf{0.69} \\
      & TCN  & 1.36* & \textbf{0.72} \\ \hline\hline
  \end{tabular}
\end{table}
Figure~\ref{fig:ventil_plots} illustrates the prediction performance of selected network architectures on a representative trajectory from the pneumatic valve dataset. 
The top two plots show the applied input voltages $u_1$ and $u_2$; the bottom plots compare the predicted control chamber pressure $p$ and the plunger position $s$ with the measured values (dashed black). 
The best-performing \textit{parallel}-trained model (p-RNN) and \textit{series-parallel}-trained model (sp-RNN) are shown. 

\begin{figure}[!h]
    \centering
    \includegraphics[width=\linewidth]{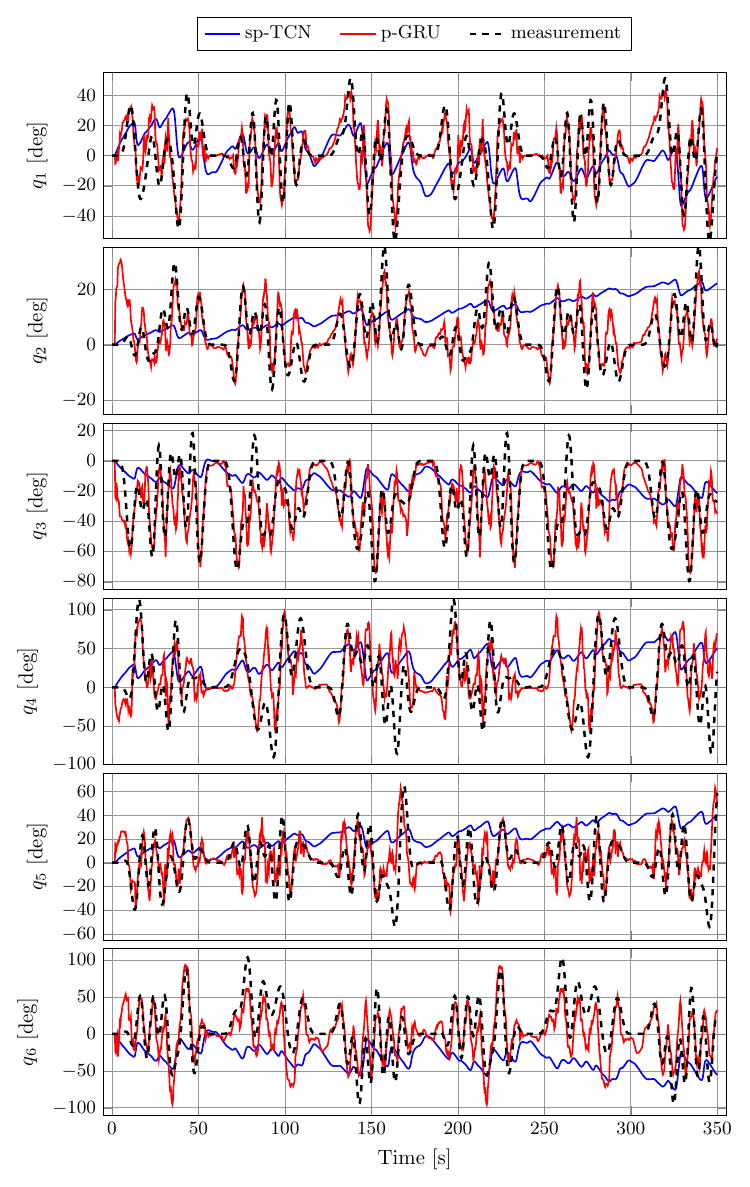}
    \caption{Predicted joint trajectories $q_1, \dots, q_6$ for the industrial robot over the full test sequence. 
    Measured positions (dashed black lines) are compared against the outputs of the best-performing \textit{parallel}-trained model (p-RNN) and \textit{series-parallel}-trained model (sp-TCN), as identified in Tab.~\ref{tab:nrmse_all}.
    Note that the desired trajectory repeats around \SI{180}{\second} due to the structure of the test set.}
    \label{fig:robot_plots}
\end{figure}

Figure~\ref{fig:robot_plots} illustrates the prediction performance of selected network architectures on the industrial robot benchmark dataset. 
Each subplot corresponds to one of the six robot joints, showing the measured trajectory (dashed black) alongside predictions from the best \textit{parallel}-trained model (p-GRU) and \textit{series-parallel}-trained model (sp-TCN). 
\section{Discussion of Results} \label{sec:discussion}
For both the pneumatic valve and the industrial robot datasets, models trained using the \textit{parallel} training strategy consistently outperformed their \textit{series-parallel} counterparts across all tested architectures.
For the pneumatic valve, Tab.~\ref{tab:nrmse_all} shows that \textit{parallel} training leads to significantly lower NRMSE across all models. 
This is also reflected in the trajectory comparisons shown in Fig.~\ref{fig:ventil_plots}.
While both strategies result in reasonably accurate predictions, networks trained in parallel are better at capturing long-term dynamics. 
In particular, \textit{series-parallel}-trained networks struggle with abrupt state transitions, such as the ones around \SI{50}{\milli\second} and \SI{100}{\milli\second}, where they deviate from the measured trajectory. 
In contrast, \textit{parallel}-trained networks maintain close alignment with the measured outputs across the entire sequence, except for a minor error at approximately \SI{40}{\milli\second}. 
This deviation corresponds to a region where the physical system transitions between valve states (open vs. closed), influenced by contact forces that were underrepresented in the training data. 
As expected, such rare behaviors are difficult for a neural model to generalize accurately.

Table~\ref{tab:nrmse_all} also shows similar trends for the industrial robot benchmark. 
\textit{Parallel}-trained models outperform their \textit{series-parallel} counterparts on all architectures, and only the \textit{parallel} models achieve NRMSE values below the baseline linear model (0.82) reported in~\cite{Weigand2022}. 
Figure~\ref{fig:robot_plots} illustrates the predicted joint positions over the full test trajectory. 
While none of the models fully captures all joint behaviors, even the best \textit{series-parallel}-trained network (sp-TCN) shows significant drift and accumulation of error. 
In contrast, \textit{parallel}-trained models track the measurements more closely and maintain better stability over long prediction horizons.

These results suggest that \textit{parallel} training is not only more robust to long-term prediction errors but also better aligned with inference-time usage, where past model outputs (not ground-truth data) are used recursively. 
This alignment appears to reduce distributional shift between training and deployment, thereby improving generalization. 
The consistent performance gains across two very different systems and five network architectures further support the recommendation that \textit{parallel} training should be adopted as the default strategy for simulation-oriented modeling tasks.

\section{Conclusion \& Outlook} \label{sec:conclusion}
This paper presented a comparative study of the two most widely used training strategies for neural network modeling of dynamical systems: \textit{series-parallel} and \textit{parallel} training. 
To provide clarity in the often inconsistent terminology found in the literature, standard terms based on a review of prior work were adopted in this work.  
The results, obtained across two real-world datasets and five neural network architectures, consistently demonstrate that \textit{parallel} training yields superior long-term predictive accuracy. 
Notably, this advantage holds independent of the specific model architecture.

The experiments relied exclusively on measurement data, underscoring the relevance of the findings to practical, data-driven modeling scenarios in control and robotics. 
In contrast to the prevailing assumption that \textit{series-parallel} training is more stable or easier to train, the findings support the use of \textit{parallel} training as the default paradigm when the goal is multi-step simulation or forecasting.

Future work will explore the application of \textit{parallel} training to more advanced and specialized neural architectures, including hybrid models or physics-informed networks. Additionally, improvements to training strategies - such as curriculum learning, online adaptation, or physics-informed loss functions - may further enhance performance and robustness. 
Another promising direction lies in formalizing the connection between training strategies and system identification theory, particularly in the context of learning-based control, e.g., learning-based MPC.

\bibliographystyle{IEEEtran}
\bibliography{library}

\end{document}